\pdfoutput=1
\documentclass[10pt,twocolumn,letterpaper]{article}
\usepackage[pagenumbers]{cvpr}

\usepackage{tabularx, array}
\graphicspath{{media/}}
\definecolor{cvprblue}{rgb}{0.21,0.49,0.74}
\usepackage[pagebackref,breaklinks,colorlinks,allcolors=cvprblue]{hyperref}
\newcommand{\rmin}{RVLM-X\textsubscript{min}}
\renewcommand{\paragraph}[1]{\vspace{.5em}\noindent\textbf{#1}}%

\title{RVLM-X: Towards a Universal VLM Harness}
\author{Aadi Palnitkar\\University of Maryland\\{\tt\small apalnitk@umd.edu}
\and Dmytro Kurdydyk\\Davidson College\\{\tt\small dmkurdydyk@davidson.edu}
\and Xiaomin Lin\\University of South Florida\\{\tt\small xlin2@usf.edu}}

\begin{document}
\maketitle
{\let\thefootnote\relax\footnotetext{A short, preliminary version of this work was presented at the LEAF Workshop at IEEE MILCOM 2026.}}
\begin{abstract}
Harnesses for vision-language models (VLMs) have grown with the models' capabilities, notably for images far larger than those seen in training, where downscaling loses small details. Most of them wrap the VLM in a fixed procedure, such as a zoom tree, that decides in advance where it looks and how the evidence is combined. These procedures make up for the weaknesses of older models by keeping them on a task-specific path, but with frontier models that reliably plan and use tools, they can lower accuracy. We propose RVLM-X, a task-generalizable recursive VLM harness that runs unchanged across photographs, charts and maps. Following the recursive inference of Recursive Language Models, the VLM programmatically controls its own visual context in a code REPL that holds the full-resolution image and calls itself recursively on regions that it chooses. With a cost-efficient frontier model as the base of every method, RVLM-X is more accurate than a single call on all 7 public benchmarks and has the highest pooled accuracy. Because existing benchmarks mostly test whether a model finds one small detail, we also introduce MapGenBench, procedurally generated maps whose questions combine fine-detail perception with multi-step reasoning. There, RVLM-X answers 71.4\% of held-out questions against 55.5\% for the strongest baseline. Overall, our findings suggest that as VLMs plan and use tools more reliably, letting the model write its own visual search works better than fixing one in advance.
\end{abstract}

\begin{figure*}[t]
  \centering
  \includegraphics[width=\textwidth]{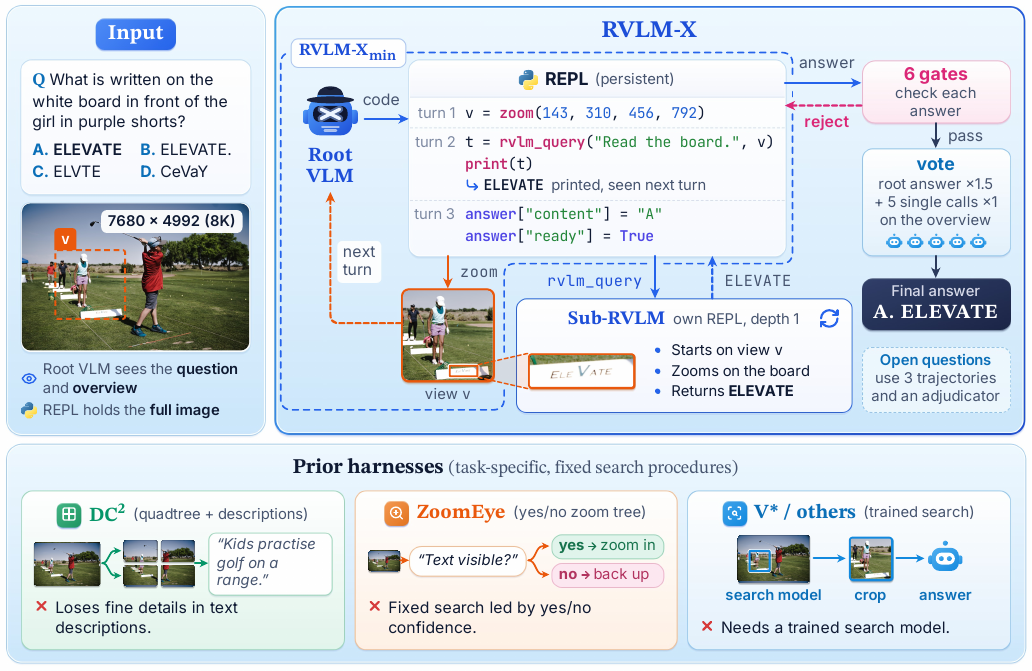}
  \caption{\textbf{RVLM-X.} The full-resolution image stays in a persistent REPL, outside the root VLM's context. The dashed outline marks \rmin{}, which runs with only \texttt{zoom}, \texttt{crop} and \texttt{show}, and Section~\ref{sec:full} describes what RVLM-X adds. In this schematic HR-Bench 8K example, the root VLM zooms in on the girl and the board, and a sub-RVLM zooms in on the board and returns ``ELEVATE''. The lower panel shows the fixed search procedures of earlier harnesses.}
  \label{fig:system}
\end{figure*}

\section{Introduction}
\label{sec:intro}
Vision-language models (VLMs) are now applied to inputs where the decisive evidence is small: a label in a high-resolution street scene \cite{wu2024vstar,wang2025dc2,zhang2025mmerealworld}, a legend entry on a topographic map \cite{pyo2026frieda,chen2026omnimapbench}, a value in a dense chart. VLMs also read documents \cite{mathew2021docvqa}, infographics \cite{mathew2022infographicvqa}, charts \cite{masry2022chartqa}, screens \cite{li2025screenspotpro}, remote-sensing \cite{kuckreja2024geochat,wang2025xlrsbench} and medical images \cite{li2023llavamed}. Because such images are far larger than those that VLMs are trained on, they are either downscaled, losing details only a few pixels high, or sent at full size, where the model does worse (Section~\ref{sec:results}). Dynamic-resolution \cite{wang2024qwen2vl,bai2025qwen25vl} and tiling encoders \cite{chen2024internvl25,li2025llavaonevision} only raise this limit, and many questions need several such details from different places in the image. Long text exceeds a language model's context window in a similar way, but here the difficulty is mainly visual.

A large image offers a simple solution, because the model can zoom into a region and look at it again at full resolution. Several harnesses build on this idea, each with a search procedure of its own. V$^*$ adds a trained search model \cite{wu2024vstar}, DC$^2$ describes a quadtree of patches in text and combines the descriptions \cite{wang2025dc2}, ZoomEye walks a tree of zoomed sub-regions guided by the model's yes/no confidences \cite{shen2025zoomeye}, and Zoom-Refine zooms once into a box that the model names and then refines its answer \cite{yu2025zoomrefine}. Other methods crop from attention maps \cite{zhang2025mllmsknow}, retrieve crops relevant to the question \cite{wang2025rap} or search with a detection model \cite{li2025dyfo,ma2026hee}, and GIANT navigates gigapixel pathology slides through crops at several magnifications \cite{buckley2025giant}. Each of these harnesses fixes in advance how the model searches and how the evidence is combined, and each depends on an assumption about the task or the model: V$^*$ requires a trained search model, ZoomEye relies on the model's yes/no probabilities, and DC$^2$ passes full-resolution detail to its answer only as text descriptions of patches. We call such harnesses \emph{task-specific}, because a designer writes their search procedure in advance, and a harness that fixes no search procedure and lets the model write it \emph{task-generalizable}. Recent work on LLM agents argues that a task-specific harness works well on the task it was built for and becomes rigid when a task breaks its assumptions \cite{li2026jaz}. Many of these harnesses also need model internals that API models hide or expose only in part, such as attention maps or token probabilities, or a trained model for search or cropping \cite{wu2024vstar,gong2026pixeleyes,li2026insighto3}, which limits them to particular models.

We base our harness, RVLM-X, on the hypothesis that stronger models, which plan and write code reliably, need fewer constraints on where to look but still gain from structure that spends more compute on the same problem, such as tools, checks and votes (Section~\ref{sec:full}). Several results on LLM agents support the first part, although not in every setting \cite{starace2026scaffold}. When models were weaker, a fixed procedure of localization, repair and validation beat open-source LLM agents on software engineering tasks \cite{xia2025agentless}. With frontier models, elaborate harnesses for ML engineering give no advantage over a minimal agent, although weaker models still benefit from them \cite{brilliantov2026harness}. Likewise, a prescribed optimization pipeline wins with a medium-sized model while open-ended optimization wins with a frontier one \cite{xue2026rethinking}, and chain-of-thought prompting adds little for models that already reason \cite{meincke2025prompting}. Closest to our setting, JAZ shows that a minimal harness, in which the model calls itself recursively and sees its own inputs and history as program variables, outperforms harnesses built specifically for memory and self-improvement \cite{li2026jaz}. Such comparisons have so far been made mainly for LLM agents on text, and we make one on vision, with the same base model for every method and across 6 base models.

\begin{figure*}[t]
  \centering
  \includegraphics[width=\textwidth]{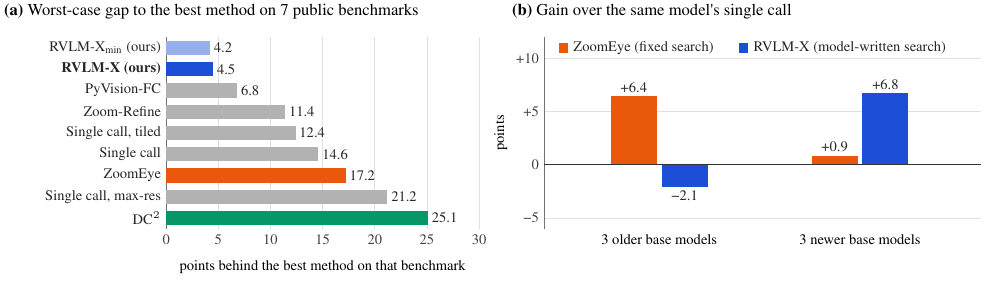}
  \caption{\textbf{(a)} Each method's largest gap, with GPT-6-luna, to the best of all methods on any of the 7 public benchmarks. \textbf{(b)} Gain over the same model's single call on 300 questions, averaged over the 3 older models and over the 3 newer ones, whose code fails in at most 8.3\% of turns, with ZoomEye's stated confidences for GPT-5.4-nano and GPT-6-luna. Figure~\ref{fig:crossover} shows each model (Section~\ref{sec:crossover}).}
  \label{fig:teaser}
\end{figure*}

RVLM-X turns the VLM into a recursive VLM (RVLM) for any task (X) by applying the method of Recursive Language Models (RLMs) \cite{zhang2025rlm}. An RLM keeps its long input, such as a set of documents, as a variable in a read-eval-print loop (REPL) instead of reading it into its context window, and the model inspects it with code and hands parts of it to recursive sub-calls. RVLM-X keeps the full-resolution image outside the model in the same way and lets the model decide where to look. Because the harness fixes no search procedure and its checks look only at the submitted answer and the evidence behind it, one configuration runs unchanged on photographs, charts and maps, and on every question format from multiple choice to free-form text. It needs no training and reads no model internals, so any model that can write code can run it. We also propose a minimal variant, \rmin, in which the model only zooms, crops and shows views (Section~\ref{sec:min}). It is far cheaper, and it separates the REPL loop itself from what RVLM-X adds to it. Figure~\ref{fig:teaser} summarizes our results.

We contribute the following:
\begin{enumerate}
\item \textbf{RVLM-X}, a training-free recursive harness that runs unchanged across benchmarks and question formats, and its minimal variant \rmin, a loop that costs a fifth as much per map question and does not differ significantly from it when an answer turns on one detail.
\item \textbf{An evaluation with the same base model for all methods} on 7 public benchmarks against 7 baselines, in which RVLM-X has the highest pooled accuracy.
\item \textbf{MapGenBench} (Section~\ref{sec:mapgenbench}), a procedurally generated map benchmark whose questions combine several small details, with a held-out split of 1,000 questions on unseen maps, where RVLM-X leads every other method by 15.9 points or more.
\item \textbf{A comparison across 6 base models}, in which ZoomEye's fixed search loses most of its gain as models improve, while RVLM-X gains once the model runs its code reliably (Section~\ref{sec:crossover}).
\end{enumerate}

\section{Related work}
\label{sec:related}

\paragraph{Model-written procedures.}
VisProg and ViperGPT generate one program over vision modules \cite{gupta2023visprog,suris2023vipergpt}, and MM-REACT and Visual ChatGPT let a text-only model call vision experts \cite{yang2023mmreact,wu2023visualchatgpt}. Visual Sketchpad, PyVision and VLM-in-Sandbox let a model write code over several turns without training \cite{hu2024sketchpad,zhao2025pyvision,yang2026vlmsandbox}. Models such as OpenAI's o3 crop and zoom in their chain of thought or run a native code interpreter \cite{openai2025thinkingimages}, and open models are trained to do the same \cite{qi2025cogcom,zheng2026deepeyes,su2025pixelreasoner}, mostly with reinforcement learning \cite{lai2026minio3,zhang2026thyme,hong2026deepeyesv2} (see also \cite{guo2026codevision,li2026s1vl,su2025thinkingimages}). \rmin{} belongs with these training-free loops but keeps only 3 tools and 1 trajectory and thus shows what the loop achieves alone, as removing sub-calls does for the RLM REPL \cite{zhang2025rlm}. RVLM-X differs from these systems, which either need training or lack recursive calls over image regions.

\paragraph{Recursive language models and agent harnesses.}
Besides RLMs \cite{zhang2025rlm}, other harnesses interleave reasoning with tool calls or code \cite{yao2023react,wang2024codeact}, split long tasks by decomposition \cite{khot2023decomposed,lee2023recursion}, threads \cite{schroeder2025thread}, memory \cite{packer2023memgpt} or chains of agents \cite{zhang2024chainofagents}, make each recursive unit a full agent \cite{lumer2026rah} or make the recursive call the single primitive of a minimal harness \cite{li2026jaz}. For vision, ReVisionLLM and sCoT are trained to search or delegate recursively \cite{hannan2025revisionllm,yang2025scot}.

\paragraph{The RVLM of Mayumu et al.}
Despite the similar name, RVLM-X differs from the training-free RVLM of Mayumu et al.\ \cite{mayumu2026rvlm}. Their model sees the image only in its first message and then reasons over text, while RVLM-X's root VLM keeps a downscaled overview and sees every view that it requests. Their sub-calls are stateless completions, while a sub-RVLM runs its own REPL loop on a view. RVLM-X also adds measuring tools, gates and a vote, and we evaluate it against matched baselines on 8 test sets and 6 base models, while their RVLM is evaluated in case studies on brain MRI and chest X-ray reports.

\paragraph{High-resolution and map benchmarks.}
HR-Bench \cite{wang2025dc2}, V$^*$Bench \cite{wu2024vstar}, VisualProbe \cite{lai2026minio3}, MME-RealWorld \cite{zhang2025mmerealworld} and O3-Bench \cite{li2026insighto3} test fine detail in large images, and FRIEDA \cite{pyo2026frieda}, OmniMapBench \cite{chen2026omnimapbench}, MapQA \cite{chang2022mapqa}, MapBench \cite{xing2025mapbench}, ReasonMap \cite{feng2026reasonmap} and ReasonMap-Plus \cite{feng2026rewardmap} test map reading. Questions in HR-Bench, V$^*$Bench, VisualProbe and MME-RealWorld mostly turn on a single detail. CLEVR, DynaMath and MapReason-OSM generate questions with exact answers procedurally \cite{johnson2017clevr,zou2025dynamath,venkatanarayanan2026mapreasonosm}, and MapGenBench (Section~\ref{sec:mapgenbench}) also generates the map geography, with deciding details a few pixels high in the overview.

\section{Method}
\label{sec:method}

As an RLM does with a long text, RVLM-X holds the full-resolution image as a variable in a persistent Python REPL. The \emph{root VLM}, the VLM that receives the question, inspects the image with code and sees it only through an overview and the views that it requests (Figure~\ref{fig:system}), so the full-resolution image stays outside its context. \rmin{} runs this loop with tools that only view the image, and RVLM-X adds the capabilities and checks of Section~\ref{sec:full}.

\subsection{\rmin: the model writes the search}
\label{sec:min}
\rmin{} leaves the search to the model, which decides what to view, in which order, at which scale and when to stop. Each turn, the root VLM returns its reasoning and a block of Python code, the code runs, and the next observation shows the printed output, any errors and the requested views. The root VLM keeps an overview in its context, the image downscaled to a long edge of at most 3840\,px and an area of at most 3840$\times$2160. To look closer, it has 3 tools: \texttt{zoom} and \texttt{crop} render a region of the source image as a new view of at most 2048\,px (upscaled at most 3$\times$), and \texttt{show} displays a view. Boxes use 0--1000 coordinates of their view, which lets the model zoom into a view again. \rmin{} runs 1 \emph{trajectory}, a run of this loop, and its first submitted answer is final. Its system prompt documents the tools and adds the 4 map rules of RVLM-X (Section~\ref{sec:full}), so that the 2 variants differ only in what the harness adds.

\subsection{RVLM-X: capabilities and checks}
\label{sec:full}
\begin{figure*}[t]
  \centering
  \includegraphics[width=\textwidth]{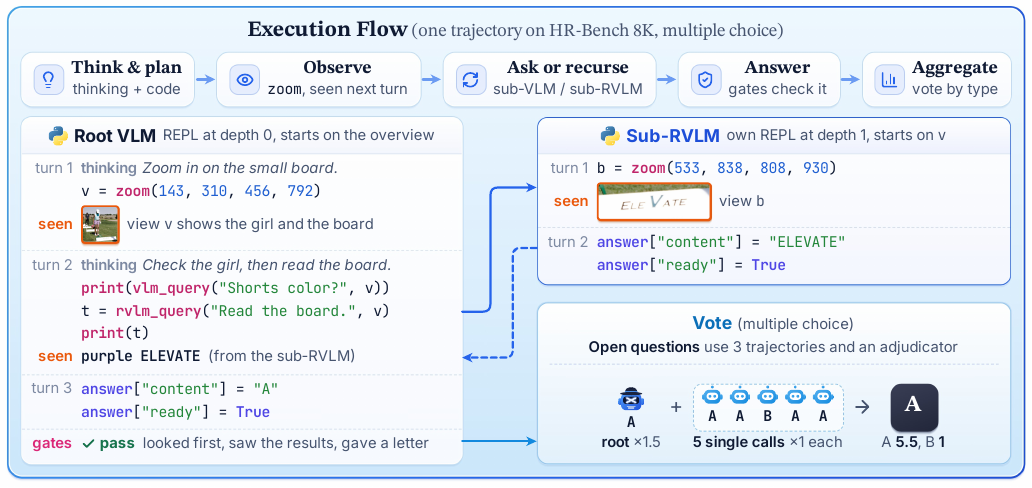}
  \caption{\textbf{One trajectory of RVLM-X, turn by turn} (HR-Bench 8K, multiple choice). Each turn is thinking plus code, whose printed output and views the root VLM sees on the next turn. In turn 2, a sub-RVLM starts on view \texttt{v}, zooms in on the board and returns ``ELEVATE''. The submitted answer passes the gates, and its letter (weight 1.5) is combined with 5 single calls on the overview (schematic).}
  \label{fig:method}
\end{figure*}

RVLM-X adds capabilities and checks to this loop but leaves the search to the model, as no part of it chooses which regions the model views or in which order. Figure~\ref{fig:method} follows one of its trajectories turn by turn.

RVLM-X adds these capabilities and checks for 3 reasons: accuracy rises with test-time compute, some failures are simple to prevent, and tools save calls. First and most important, reasoning before answering \cite{wei2022cot}, sampling and comparing several answers \cite{wang2023selfconsistency,snell2025scaling}, and taking more turns \cite{lai2026minio3} all raise accuracy. RVLM-X therefore makes the model spend more compute on looking: the first turn asks it to plan and inspect before it answers, the gates send it back to look again, sub-RVLMs give a subtask its own loop, and the vote weighs several attempts. Second, failures such as answering from the overview, before seeing the requested views or in the wrong format are simple, and a fixed rule or check prevents them cheaply. Third, tools take over work that would otherwise consume a call, such as estimating a distance or matching a color by eye, so that each call can spend its reasoning on the question itself.

\paragraph{Prompt and first turn.}
The system prompt documents all tools and gives 12 working rules, such as to restate what the answer must contain, to zoom in before concluding and to enumerate every candidate for questions about all, which or how many. Four of them say how to read maps (legends, north arrows and scale bars; features versus labels; spatial relations; color ramps), and we wrote them on 20 FRIEDA questions, so FRIEDA is in-distribution for the prompt. The first turn's instruction asks the root VLM to restate the ask, note a first impression and inspect the deciding details at full resolution before it submits.

\paragraph{Tools and sub-RVLMs.}
Besides \texttt{zoom}, \texttt{crop} and \texttt{show}, RVLM-X has helpers, measuring tools and sub-RVLM calls. As helpers, \texttt{tiles} splits a view into a grid, a sub-VLM call (\texttt{vlm\_query}) asks the VLM one question about a view, and \texttt{locate} returns boxes and points for named objects from one VLM call. The measuring tools \texttt{to\_pixels}, \texttt{dist}, \texttt{bearing}, \texttt{color\_at} and \texttt{colors} measure on the source pixels, and \texttt{mark} numbers points on a view. A sub-RVLM call (\texttt{rvlm\_query}) delegates a subtask to a new REPL loop on a given view, for at most 5 turns and at depths 1 and 2, and returns only its answer, which keeps the subtask's views out of the root's context. All calls of a question share a budget of 60.

\paragraph{Gates, vote and answer repair.}
Every submitted answer goes through 6 gates, and a failed gate returns its reason on the next turn. The first 5 reject (i) an empty answer; (ii) an answer given before the root VLM has looked beyond the overview; (iii) an answer given in the same turn as views that it has not yet seen; (iv) a refusal or a claim that the answer cannot be determined; and (v) an answer in the wrong format. The sixth, a premise check on open questions, asks once per trajectory for the evidence behind each condition. A trajectory ends when an answer passes or after 8 turns (10 on multiple choice, where only 1 trajectory runs). On open questions, 3 trajectories vote as in self-consistency \cite{wang2023selfconsistency}, and an adjudicator that sees their last views decides when no 2 agree \cite{chen2023usc}. On multiple choice, where single calls on the overview already answer most questions, the root's letter (weight 1.5) is combined with those of 5 single calls, plain calls of the base model on the overview. Answer repair rewrites a final answer that still fails the format, or answers the question itself when the final text is not an answer at all.

\section{MapGenBench}
\label{sec:mapgenbench}

MapGenBench tests whether a method can find several small details in a large image and combine them. A procedural generator draws a 10240$\times$5760\,px national-park map from a seed, with about 200 features such as zones, roads, visitor facilities, checkpoints and campsites, and stores the exact geometry of every feature, which makes the questions generated from it correct by construction. The deciding codes are about 9\,px tall at full resolution and 3.4\,px in the 3840$\times$2160 overview.

The questions come in equal numbers from 5 task families: distance and direction, route tracing, conditional counts, false premises and multi-region questions. \emph{Distance and direction} asks for the distance or compass direction between 2 features in different quadrants, named only by their codes (``What is the straight-line distance between campsite C-613 and visitor facility F-412, in kilometres?'', 92.0\,km). \emph{Route tracing} asks for the $N$th checkpoint along a long route that crosses others (``Following Komkrev Road from its western end, what is the code of the ninth checkpoint you pass?'', P7X-76). \emph{Conditional counts} ask how many facilities whose ID meets a condition lie inside a dense zone, with 5--6 of them a few pixels from its outline (``How many visitor facilities whose ID number is less than 400 are inside the Griloum Uplands zone?'', 7). \emph{False premises} name a look-alike of a real code, such as F-483 on a map that has F-283 (``not on the map''), or name a real code, so ``not on the map'' is not always right. \emph{Multi-region} questions ask which of 3--5 zones holds the most facilities that meet a condition (``Which of the Sastir Basin, Bregov Slopes and Drabrail Basin zones contains the most visitor facilities whose ID ends in an even digit?'', Sastir Basin). Fixed rules score the reply: a distance within 10\%, the exact compass sector, count, code or zone, or ``not on the map''.

The single call is a weak reference here, because it cannot read the deciding codes, which its overview shrinks to 3.4\,px, and it answers 19.7\% of the held-out questions. A stronger reference is a single call that also gets full-resolution tiles and sees every label, but it must still find the few deciding features among about 200.

\paragraph{Held-out split.}
We use 1,000 questions on 100 maps, fixed in advance and never used to build RVLM-X.

\section{Experiments}
\label{sec:experiments}

\begin{table*}[t]
  \caption{Accuracy (\%) with GPT-6-luna as the base model of every method, grouped as in Section~\ref{sec:setup}. \emph{Total avg.}\ is the pooled accuracy over all 7,223 questions, and FRIEDA is in-distribution for the prompts of RVLM-X and \rmin. The single-call row is the better of 2 runs on each public benchmark. Bold marks the best in each column.}
  \label{tab:main}
  \centering
  \footnotesize
  \setlength{\tabcolsep}{5pt}
  \begin{tabular}{@{}lrrrrrrrr@{}}
    \toprule
    Method & HR-Bench & FRIEDA & O3-Bench & VisualProbe & MME-RW-Lite & OmniMapBench & ReasonMap-Plus & Total avg. \\
    \midrule
    Single call                               & 89.8 & 58.2 & 81.4 & 53.6 & 62.6 & 74.9 & 94.3 & 74.0 \\
    Single call, max-res            & 78.5 & 56.0 & 77.7 & 47.0 & 61.0 & 71.8 & 93.0 & 71.0 \\
    Single call, tiled              & 90.8 & 56.8 & \textbf{85.2} & 57.3 & 62.9 & 74.0 & 95.0 & 74.4 \\
    \midrule
    DC$^2$ \cite{wang2025dc2}                 & 85.0 & 49.0 & 72.8 & 43.1 & 55.5 & 69.0 & 92.1 & 67.9 \\
    ZoomEye \cite{shen2025zoomeye}            & 88.8 & 52.0 & 79.4 & 55.5 & 62.0 & 74.5 & 93.0 & 73.0 \\
    Zoom-Refine \cite{yu2025zoomrefine} & \textbf{92.0} & 57.8 & 84.9 & 61.7 & 63.8 & 77.0 & 96.5 & 76.2 \\
    \midrule
    PyVision-FC \cite{zhao2025pyvision} & 90.0 & 64.0 & 83.8 & 61.4 & 62.3 & \textbf{78.4} & 95.4 & 76.2 \\
    \midrule
    \rmin                           & \textbf{92.0} & 65.0 & 81.4 & \textbf{68.2} & 63.2 & 76.0 & 94.7 & 76.2 \\
    RVLM-X                                    & 91.0 & \textbf{69.2} & 84.9 & 63.7 & \textbf{63.9} & 77.9 & \textbf{97.3} & \textbf{77.5} \\
    \bottomrule
  \end{tabular}
\end{table*}

\subsection{Experimental setup}
\label{sec:setup}

\paragraph{Base models and benchmarks.}
GPT-6-luna is the base model of every method and the judge where a scoring rule needs one. Section~\ref{sec:crossover} adds InternVL2.5-8B \cite{chen2024internvl25} and Qwen2.5-VL-7B \cite{bai2025qwen25vl}, served locally with vLLM, and GPT-4.1-nano, GPT-5.4-nano and GPT-5.6-luna. Methods reason whenever the base model can, and with GPT-5.4-nano every method runs at medium effort. Our 8 test sets are 7 public benchmarks with 7,223 questions and held-out MapGenBench. Two public benchmarks are mostly on photographs, HR-Bench 4K and 8K (400) \cite{wang2025dc2} and VisualProbe (515) \cite{lai2026minio3}. MME-RealWorld-Lite (MME-RW-Lite, 1,919) \cite{zhang2025mmerealworld} covers text in the wild, remote sensing, diagrams and tables, monitoring and driving, and O3-Bench (345) \cite{li2026insighto3} covers composite charts and maps. The other 3 are on maps: FRIEDA (500, direct setting) \cite{pyo2026frieda}, OmniMapBench (2,096) \cite{chen2026omnimapbench} and the test split of ReasonMap-Plus (1,448 on 11 transit maps) \cite{feng2026rewardmap}.

\paragraph{Resolution.}
All methods use the API's default image handling, as no call sets the API's \texttt{detail} field. With this default, the API does not downsample and rejects images above about 31\,MP (30,000 patches of 32$\times$32\,px). The single call receives the same overview as the root VLM (Section~\ref{sec:min}) and thus starts from the same pixels. This overview is a resized image on every HR-Bench and MapGenBench question and on 40 to 486 questions of each of FRIEDA, O3-Bench, VisualProbe, MME-RW-Lite and OmniMapBench, and on 837 of ReasonMap-Plus.

\paragraph{Baselines.}
We compare with external methods in 3 groups: one call, designer-written procedures and model-written procedures. \emph{One call}: the single call, a \emph{max-resolution} single call with up to 29,000 patches, and a \emph{tiled} single call that adds full-resolution tiles of at most 2048\,px to the overview, each captioned with its box. \emph{Designer-written procedures}: ZoomEye, DC$^2$ and Zoom-Refine. We port ZoomEye and DC$^2$ to GPT-6-luna, keeping their prompts, thresholds and search and adapting only the parts that need internals the API hides, and on InternVL2.5-8B our ZoomEye port is within sampling error of the published HR-Bench results. \emph{Model-written procedures}: PyVision with its code tool as an API function (PyVision-FC), since its text protocol produced no code with GPT-6-luna. Because ReasonMap-Plus, like MapGenBench, asks for answers that combine several readings of a map, every baseline except ZoomEye and DC$^2$ also receives RVLM-X's 4 map rules on both. On the other public benchmarks, the baselines run without them, since the rules change PyVision-FC and \rmin{} by under 1 point pooled.

\paragraph{Statistics and variance.}
Differences are paired over questions and computed before rounding, with 95\% intervals from a paired bootstrap with 10,000 resamples and $p$-values from an exact two-sided sign test, and on held-out MapGenBench a second bootstrap resamples whole maps. A GPT-6-luna judge scores the free-form answers of FRIEDA and VisualProbe under each benchmark's rules, option letters and set rules score HR-Bench, O3-Bench, MME-RW-Lite and OmniMapBench, exact match scores ReasonMap-Plus, and MapGenBench uses its fixed rules. An API error counts as a wrong answer, and no run had one. For paired tests against the max-res and tiled single calls, Zoom-Refine, PyVision-FC and \rmin, RVLM-X also answered 1,415 questions outside ReasonMap-Plus: all of FRIEDA and VisualProbe and 100 from each of the other 4. There, it answers 67.4\% and 63.3\% of FRIEDA and VisualProbe, against 69.2\% and 63.7\% in Table~\ref{tab:main}.

\subsection{Comparison with existing harnesses}
\label{sec:results}

\begin{table}[t]
  \caption{Accuracy (\%) on held-out MapGenBench with GPT-6-luna, overall and per family (200 questions each). \$/q is the cost and Calls the model calls per question, and bold marks the best accuracy in each column.}
  \label{tab:maps}
  \centering
  \footnotesize
  \setlength{\tabcolsep}{1.7pt}
  \begin{tabular}{@{}lrrrrrrrr@{}}
    \toprule
     & \multicolumn{6}{c}{MapGenBench, held-out} & & \\
    \cmidrule(lr){2-7}
    Method & All & Count & Dist. & Multi & Prem. & Route & \$/q & Calls \\
    \midrule
    Single call   & 19.7 & 10.5 & 43.0 & 26.0 & 8.5  & 10.5 & 0.002 & 1 \\
    Single call, tiled & 55.5 & 33.5 & 80.0 & 50.5 & \textbf{71.5} & 42.0 & 0.004 & 1 \\
    ZoomEye       & 20.9 & 14.0 & 41.0 & 26.5 & 11.0 & 12.0 & 0.014 & 57.9 \\
    DC$^2$        & 19.9 & 8.0  & 16.5 & 32.5 & 37.0 & 5.5  & 0.006 & 21.7 \\
    Zoom-Refine   & 32.9 & 45.5 & 42.0 & 21.0 & 36.5 & 19.5 & 0.004 & 2 \\
    PyVision-FC   & 48.0 & 49.5 & 39.5 & 57.5 & 49.0 & 44.5 & 0.014 & 11.3 \\
    \rmin         & 53.7 & 52.0 & 62.0 & 66.0 & 52.0 & 36.5 & 0.007 & 5.4 \\
    RVLM-X        & \textbf{71.4} & \textbf{78.0} & \textbf{90.5} & \textbf{84.0} & 56.0 & \textbf{48.5} & 0.034 & 29.7 \\
    \bottomrule
  \end{tabular}
\end{table}

\textbf{\textit{On every public benchmark, RVLM-X is within 4.5 points of the best method.}}
Because RVLM-X runs unchanged across tasks, it must be accurate on every benchmark, and we therefore compare each method's largest gap to the best method on any public benchmark (Table~\ref{tab:main}, Figure~\ref{fig:teaser}a). RVLM-X and \rmin{} fall at most 4.5 and 4.2 points behind, while every other method falls 6.8 to 25.1 points behind, each time on FRIEDA or VisualProbe. Without FRIEDA, both still keep the smallest gaps.

\textbf{\textit{Outside ReasonMap-Plus, RVLM-X does not differ significantly from \rmin{} or PyVision-FC.}}
RVLM-X has the highest pooled accuracy (77.5\%) and stays highest without FRIEDA (78.1\%). On the 1,415 paired questions, the sign test gives $p{=}0.38$ against \rmin{} and $p{=}0.26$ against PyVision-FC. It is significantly ahead of Zoom-Refine there, by 4.1 points [1.9, 6.3], mostly on FRIEDA, and of the single call, by 8.6 [6.3, 10.8]. The fixed searches of DC$^2$ and ZoomEye fall below the single call overall, each where its assumption fails. DC$^2$ passes the patches to its answer only as text descriptions, which lose the deciding details, and trails the single call on all 7 benchmarks. ZoomEye relies on yes/no confidences that GPT-6-luna saturates, and many of its searches therefore stop at the top of the zoom tree.

\textbf{\textit{Sending more pixels in one call gains at most 1.0 point pooled.}}
The simplest alternative to a harness is to send more pixels in one call, as the max-resolution and tiled single calls do. Paired with a single call run alongside them, the max-resolution call loses 2.8 points outside ReasonMap-Plus [$-$3.5, $-$2.1] and 10.2 on HR-Bench, and the tiled call gains only 1.0 [0.4, 1.6]. By comparison, RVLM-X's paired run leads the tiled call by 10.6 and 6.0 points on FRIEDA and VisualProbe (both $p<0.01$).

\textbf{\textit{More calls do not explain RVLM-X's gain.}}
RVLM-X makes up to 60 calls per question, so its gain could come from the number of calls alone. To test this on the open questions of FRIEDA and VisualProbe, we let independent single calls on the overview vote, as in self-consistency \cite{wang2023selfconsistency}, with at least as many calls as RVLM-X's cost would buy (26 and 16 per question). The self-consistency vote stops improving after 3 calls, and RVLM-X stays 7.0 [3.2, 10.8] and 8.9 [4.7, 13.2] points above it, because more samples of the same overview cannot show details that the overview lacks. On multiple choice, RVLM-X itself votes with 5 single calls, and this vote could explain its gain. Over the 4,125 questions with a single correct option, however, its root trajectory alone answers 75.2\%, against 73.1\% for the single call and 74.7\% for a majority of the 5 calls. The vote raises it only to 75.6\%.

\begin{figure}[t]
  \centering
  \includegraphics[width=\columnwidth]{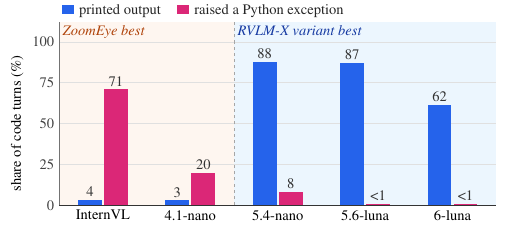}
  \caption{Share of RVLM-X's code turns that print output and that raise a Python exception, per base model. The 2 shares need not sum to 100, since a turn can also run without printing or failing, for example one that only requests views. Shading marks the best harness per model, as in Figure~\ref{fig:crossover}.}
  \label{fig:codeturns}
\end{figure}

\textbf{\textit{On MapGenBench, RVLM-X leads every other method by 15.9 points or more.}}
The public benchmarks mostly turn on a single detail, while every MapGenBench answer combines several. Here the strongest baseline is the tiled single call, which sees every label. RVLM-X leads it by 15.9 points [12.0, 19.8] and wins 4 of the 5 families (Table~\ref{tab:maps}). Unlike on the public benchmarks, \rmin{} does not beat the tiled call here, so RVLM-X's lead comes from the prompt, capabilities and checks that it adds (Section~\ref{sec:ablations}). RVLM-X loses only false premises, likely because the tiled call sees every code at once, while a search that finds no match cannot be sure the code is absent. ZoomEye and DC$^2$ gain at most 1.2 points over the single call and fall below it on ReasonMap-Plus, where RVLM-X significantly beats every baseline but Zoom-Refine.

\subsection{Older and newer base models}
\label{sec:crossover}

\begin{figure}[t]
  \centering
  \includegraphics[width=\columnwidth]{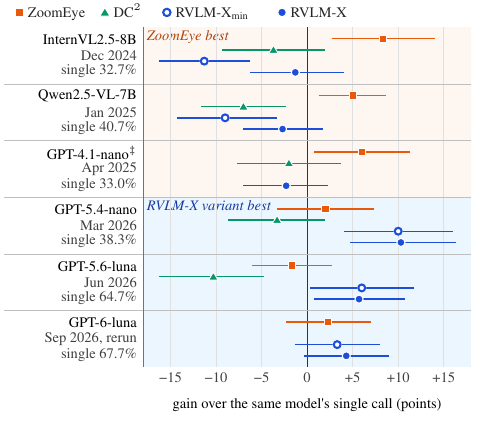}
  \caption{Gain over the same model's single call on 300 questions with 95\% intervals, for 6 base models in order of release. ZoomEye states its confidences in text with GPT-5.4-nano and in the GPT-6-luna rerun, and DC$^2$ is not shown for GPT-6-luna. Unlike in the other experiments, \rmin{} runs without the map rules, and on InternVL2.5-8B and GPT-5.6-luna it is compared with its own single call. $^\ddagger$ 1 run, without \rmin.}
  \label{fig:crossover}
\end{figure}

To test our hypothesis across base models, we ran the single call, ZoomEye, DC$^2$, \rmin{} and RVLM-X on the same 300 questions (50 HR-Bench 4K, 50 HR-Bench 8K, 100 VisualProbe, 100 FRIEDA). The 3 older base models are InternVL2.5-8B, Qwen2.5-VL-7B and GPT-4.1-nano (released by April 2025), and the 3 newer ones are GPT-5.4-nano, GPT-5.6-luna and GPT-6-luna (from March 2026). \rmin{} ran with all but GPT-4.1-nano, and Figure~\ref{fig:crossover} shows each method's gain over the same model's single call.

\textbf{\textit{Model-written search helps the models that execute its code reliably.}}
With the 3 older models, ZoomEye gains 5.0 to 8.3 points over the single call and RVLM-X loses 1.3 to 2.7. With the 3 newer models the order reverses, as RVLM-X gains 4.3 to 10.3 points and ZoomEye at most 2.3, and DC$^2$ stays below the single call with every model. The 2 groups also differ in how reliably the model's code runs, and a turn whose code fails can return no new view. With InternVL2.5-8B and GPT-4.1-nano, 71\% and 20\% of RVLM-X's code turns raise a Python exception, against at most 8.3\% with the newer models (Figure~\ref{fig:codeturns}, not measured for Qwen2.5-VL-7B). The RLM authors saw the same limit on long text, where models without enough coding ability, such as Qwen3-8B, struggled as RLMs \cite{zhang2025rlm}. RVLM-X also loses with GPT-4.1-nano, an API model, so the reversal does not follow how the model is served.

\textbf{\textit{Single-call accuracy alone does not predict whether RVLM-X helps.}}
A simpler explanation would be that RVLM-X only needs a model that is accurate enough. GPT-4.1-nano (33.0\%) and GPT-5.4-nano (38.3\%) start 5.3 points apart, but RVLM-X gains only with GPT-5.4-nano, and Qwen2.5-VL-7B starts higher (40.7\%) and still loses.

\textbf{\textit{The reversal keeps its direction under 3 controls.}}
First, GPT-5.4-nano reasons only when asked, and by default RVLM-X asks for medium effort while the single call does not, and part of RVLM-X's gain could thus come from reasoning. When every method reasons at medium effort, its gain shrinks from 15.0 to 10.3 points but stays significant ($p<0.001$), and ZoomEye gains 2.0. Second, because API log-probabilities saturate (54\% are exactly 1), we let ZoomEye state its confidences in text. Its gain with GPT-6-luna then rises from $-$0.3 to 2.3 points, not significantly below RVLM-X's 4.3 ($p{=}0.49$). Third, on the open models ZoomEye reads the token probabilities that it was designed for, and with Qwen2.5-VL-7B it gains 5.0 points ($p{=}0.017$) while RVLM-X loses 2.7, which shows that ZoomEye's lead does not depend on API access.

\textbf{\textit{\rmin{} shows the same reversal, which points to the loop itself.}}
\rmin{} gains 3.3 to 10.0 points with the newer models and loses 11.3 and 9.0 with the 2 open older models. With Qwen2.5-VL-7B, RVLM-X leads \rmin{} by 6.4 points on 299 paired questions ($p{=}0.03$), likely because its vote, gates and answer repair recover answers that a failing root trajectory loses, but it still loses 2.7 points to the single call.

\subsection{What RVLM-X adds to \rmin}
\label{sec:ablations}

\begin{table}[t]
  \caption{From \rmin{} to RVLM-X on the 1,000 held-out MapGenBench questions with GPT-6-luna, 1 run per variant. The middle row is RVLM-X without its gates, vote, sub-RVLMs and measuring tools. \emph{Step} is the gain over the row above, with the sign test, and the lower block gives the accuracy per family.}
  \label{tab:buildup}
  \centering
  \footnotesize
  \setlength{\tabcolsep}{3.2pt}
  \begin{tabular}{@{}lrrrrr@{}}
    \toprule
    Variant & Acc. & Step & $p$ & Calls & \$/q \\
    \midrule
    \rmin                        & 53.7 & --    & --                 & 5.4  & 0.007 \\
    + prompt, helpers, first turn  & 64.4 & +10.7 & $3{\times}10^{-9}$ & 9.5  & 0.011 \\
    + 4 components (RVLM-X)      & 71.4 & +7.0  & $9{\times}10^{-6}$ & 29.7 & 0.034 \\
    \midrule
     & Count & Dist. & Multi & Prem. & Route \\
    \midrule
    \rmin                        & 52.0 & 62.0 & 66.0 & 52.0 & 36.5 \\
    + prompt, helpers, first turn  & 61.0 & 79.0 & 76.5 & 56.5 & 49.0 \\
    + 4 components (RVLM-X)      & 78.0 & 90.5 & 84.0 & 56.0 & 48.5 \\
    \bottomrule
  \end{tabular}
\end{table}

\textbf{\textit{On held-out MapGenBench, both steps from \rmin{} to RVLM-X raise accuracy, the second where an answer needs an exact count or a measurement.}}
To see which parts of RVLM-X explain its lead over \rmin{} on MapGenBench, we added them in 2 steps (Table~\ref{tab:buildup}). The full prompt, the helpers and the first turn's instruction add 10.7 points [7.3, 14.2], and this middle variant is 8.9 points above the tiled single call ($p{=}10^{-5}$). The gates, vote, sub-RVLMs and measuring tools add 7.0 more [4.0, 10.0] at 3 times the cost. This gain lies mostly in conditional counts (+17.0) and distance and direction (+11.5), the 2 families that need an exact count in one zone or a measurement on the source pixels, while false premises and route tracing gain nothing.

\textbf{\textit{Outside ReasonMap-Plus, RVLM-X does not differ significantly from \rmin.}}
On the 1,415 paired questions, it is 1.0 point behind ($p{=}0.38$). On ReasonMap-Plus, it leads by 2.6 points ($p<0.001$). The same holds for RLMs on long text, where the REPL alone sometimes scores higher than the full RLM, while recursive sub-calls help on information-dense tasks \cite{zhang2025rlm}. \rmin{} thus suits questions that turn on one detail, as on most public benchmarks, and RVLM-X questions that combine several, as on ReasonMap-Plus and MapGenBench.

\section{Conclusions}
\label{sec:conclusion}

We introduce RVLM-X, a task-generalizable harness for frontier VLMs that applies recursive inference to visual reasoning. It has the highest pooled accuracy on 7 public benchmarks, is within 4.5 points of the best method on each, and leads every other method on MapGenBench by 15.9 points or more.

As for limitations, RVLM-X brings no gain with the older models, where ZoomEye is the better harness, but ZoomEye's fixed search loses most of its gain with the newer models, while model-written search gains. RVLM-X also costs more, \$0.034 per map question against \$0.004 for the tiled single call, but it gains most where questions combine many details, and \rmin{} is the cheaper choice where an answer turns on one detail. Our main results use one base model, which also judges the free-form answers of FRIEDA and VisualProbe, while the other test sets, MapGenBench included, are scored without a judge. Compositional map reasoning is tested mainly on our synthetic maps, since the single call already scores 94.3\% on ReasonMap-Plus, but their answers are correct by construction. As VLMs plan and use tools more reliably, we expect harnesses that leave the search to the model to gain further, and we hope RVLM-X can serve as a basis for them. We will release our code, map generator and held-out questions.

\section*{LLM Usage Statement}
Large language and vision-language models are the subject of this work. RVLM-X and the baselines run on the base models named in Sections~\ref{sec:setup} and~\ref{sec:crossover}, and GPT-6-luna is also the judge where a benchmark's scoring rule needs one. We used Claude Code (Anthropic) as an assistant for coding, for computing statistics and for editing the text and figures. We defined each task and reviewed and revised its output over many iterations. We conceived the method, the benchmark and the experiments, made all design decisions, and verified all code, results and text.

\end{document}